\documentclass[preprint,12pt, authoryear]{elsarticle}

\usepackage{amssymb}
\usepackage{amsmath}
\usepackage{xcolor}
\usepackage{float}
\usepackage[linesnumbered,ruled]{algorithm2e}
\usepackage{subcaption}
\usepackage{hyperref}
\hypersetup{
    breaklinks=true, 
    colorlinks=false, 
    urlcolor=black 
}

\journal{Engineering Applications of Artificial Intelligence }

\begin{document}

\begin{frontmatter}



\title{Uncertainty Guided Online Ensemble for Non-stationary Data Streams in Fusion Science} 


\author[1,2]{Kishansingh Rajput}
\author[1]{Malachi Schram}
\author[3]{Brian Sammuli}
\author[2]{Sen Lin}

\affiliation[1]{organization={Thomas Jefferson National Accelerator Facility},
            city={Newport News},
            postcode={23606}, 
            state={VA},
            country={USA}}
\affiliation[2]{organization={Department of Computer Science, University of Houston},
            city={Houston},
            postcode={77204},
            state={TX},
            country={USA}}
\affiliation[3]{organization={General Atomics},
            city={San Diego},
            postcode={92121},
            state={CA},
            country={USA}}

\begin{abstract}
Machine Learning (ML) is poised to play a pivotal role in the development and operation of next-generation fusion devices. Fusion data shows non-stationary behavior with distribution drifts, resulted by both experimental evolution and machine wear-and-tear. 
ML models assume stationary distribution and fail to maintain performance when encountered with such non-stationary data streams.
Online learning techniques have been leveraged in other domains, however it has been largely unexplored for fusion applications.
In this paper, we present an application of online learning to continuously adapt to drifting data stream for prediction of Toroidal Field (TF) coils deflection at the DIII-D fusion facility.
The results demonstrate that online learning is critical to maintain ML model performance and reduces error by 80\% compared to a static model.
Moreover, traditional online learning can suffer from short-term performance degradation as ground truth is not available before making the predictions. As such, we propose an uncertainty guided online ensemble method to further improve the performance. 
The Deep Gaussian Process Approximation (DGPA) technique is leveraged for calibrated uncertainty estimation and the uncertainty values are then used to guide a meta-algorithm that produces predictions based on an ensemble of learners trained on different horizon of historical data. 
The DGPA also provides uncertainty estimation along with the predictions for decision makers. 
The online ensemble and the proposed uncertainty guided online ensemble reduces predictions error by about 6\%, and 10\% respectively over standard single model based online learning.   
\end{abstract}



\begin{keyword}


Online Learning, Uncertainty Quantification, Data Drift, Fusion Science, Machine Learning, Concept Drift, Online Ensemble, TF-Coil, B-coil
\end{keyword}

\end{frontmatter}


\section{Introduction}
\label{sec:introduction}

The pursuit of controlled nuclear fusion as a viable energy source has led to the development of increasingly complex fusion facilities, such as tokamaks~\citep{GLISS2022113068}, which demand precise control, real-time monitoring, and predictive capabilities to optimize performance and ensure operational safety.
Traditional solutions in fusion research have relied heavily on physics-based modeling, often requiring high-fidelity simulations that are computationally expensive and not easily adaptable in real time. 
Consequently, data‑driven approaches — especially those built on Machine Learning (ML) — are set to become central to the design and operation of next‑generation fusion devices.
ML has been emerging as a transformative tool for fusion research and operations~\citep{10.1063/5.0273586}. 
By integrating sensor data with advanced computational models, ML enables modeling complex processes, supporting tasks such as diagnostics, forecasting, anomaly detection, and decision support.

\begin{figure}
    \centering
    \includegraphics[width=0.5\linewidth]{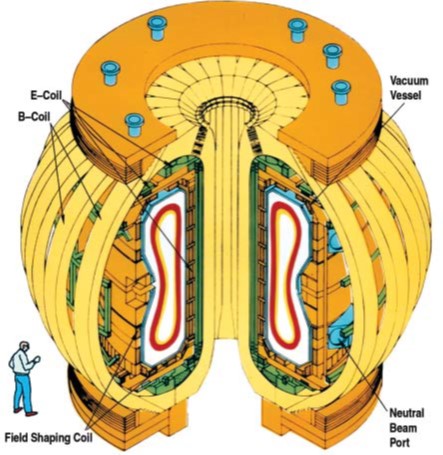}
    \caption{Sketch of DIII-D fusion tokamak}
    \label{fig:DIII-D}
\end{figure}

One particularly compelling application is the modeling and monitoring of the toroidal field (TF) coils (also called B-coils) in tokamaks. These coils are responsible for producing the strong magnetic fields required for plasma confinement, and are subject to extreme electromagnetic forces during high-field operation. 
Figure~\ref{fig:DIII-D} shows a sketch of DIII-D tokamak~\citep{d3dOverview} with position of B-coils.
At the DIII-D tokamak, calibration experiments have revealed that under certain full-field configurations, the outer legs of the TF coils experience lateral deflections large enough to cause mechanical interference and transient electrical shorts \citep{reis2003}. These deflections are sensitive to subtle mechanical slip between unbonded coil turns, thermal expansion, and the evolving electromagnetic loading profile—all of which are difficult to model analytically in real time.
Traditional finite element simulations have helped establish upper bounds on allowable lateral loads, but these models are not well suited for fast inference, adaptive prediction, or incorporation of real-time data. This opens an opportunity for ML methods—particularly those emphasizing uncertainty quantification to augment or partially replace traditional models with fast, adaptive ML models that support control-relevant decision-making.



ML typically operates under the assumption that training data is independent and identically distributed (IID) with test or inference datasets. 
However, this premise often falters in environments characterized by streaming or non-stationary data, such as fusion experiments, where data 
distributions can change over time. 
In these scenarios, new data samples may significantly differ from or even be unrelated to the original training distribution. These instances are known as out-of-distribution (OOD) samples, and ML models often struggle to make accurate predictions for them.
Deep neural networks (DNNs)~\citep{Carleo:2019ptp, RevModPhys.91.045002, Goodfellow-et-al-2016}, despite their ability to model complex systems 
effectively, face challenges in online applications in non-stationary environments.
Their limitations include: (1) producing unreliable predictions on new OOD data samples, and (2) lacking mechanisms to assess uncertainty, which makes it difficult to determine when predictions are reliable.
Recent progress in quantifying uncertainty for DNNs has introduced methods like Ensemble techniques~\citep{https://doi.org/10.48550/arxiv.2007.08792}, Bayesian Neural Networks (BNNs)~\citep{https://doi.org/10.48550/arxiv.1506.02142}, and Deep Quantile Regression (DQR)~\citep{koenker_2005}.  
These methods lack explicit distance awareness, limiting their effectiveness in capturing OOD uncertainties. 
Deep Gaussian Process Approximation (DGPA) has emerged as a promising alternative, providing single-inference predictions that are inherently capable of modeling OOD uncertainties by design~\citep{PhysRevAccelBeams.26.044602}.

Moreover, to address non-stationary data distributions, online learning~\citep{HOI2021249} has evolved as a promising technique to continuously adapt ML models with new data.
Online learning has been largely unexplored for fusion science applications.
As such, to fill this gap, we present this study to evaluate online learning algorithms for prediction of TF coils deflection at the DIII-D National Fusion Facility at General Atomics.
We use DGPA models to provide reliable uncertainty estimations along with the predictions. 
In addition, we leverage uncertainty estimation from the DGPA models to enable an uncertainty guided online ensemble to further improve the predictions. 

Our contributions are summarized in the following points.
\begin{enumerate}
    \item We present an application of online learning in Fusion applications which has been largely unexplored despite being a critical need in facilitating long term ML deployment. We present a clear comparison between static and online models performance for TF coil deflection prediction at DIII-D national fusion facility.
    \item We propose a novel uncertainty guided online ensemble method that outperforms traditional online learning and naive ensembles without uncertainty guidance.
    \item Overall, we present a case-study for a self-sustaining ML framework that provides reliable uncertainty aware predictions for Fusion applications under non-stationary data streams.
\end{enumerate}
\section{Related Work}
\label{sec:related_work}

Machine learning (ML) has become a central tool in fusion science, particularly in the prediction and mitigation of plasma disruptions. One of the landmark contributions is the work of Kates-Harbeck et al.~\citep{Kates-Harbeck2019}, who demonstrated a deep learning framework capable of predicting disruptive instabilities from multivariate diagnostic data. A key strength of their approach is cross-machine generalization: models trained on data from one tokamak (e.g., DIII-D) could be successfully applied to another (e.g., JET), highlighting the potential of ML for device-independent prediction.

Building on these advances, a recent work~\citep{https://doi.org/10.1002/ctpp.202200095} integrated a disruption predictor into a plasma control system. This work represents an important step toward real-time deployment by combining predictive scores with interpretability metrics, such as sensitivity indicators that provide insight into the causes of predicted disruptions. Such integration illustrates the feasibility of using ML models not just as offline tools, but as active components of plasma control.

Beyond deep learning architectures, statistical approaches have also contributed to disruption forecasting. Tinguely et al.~\citep{Tinguely_2019} applied survival analysis in combination with random forest models to quantify warning times and hazard functions. This probabilistic perspective provides a complementary way to evaluate the risk of disruption and offers more flexibility in characterizing temporal aspects of prediction.

Another line of research emphasizes the challenges of cross-machine and cross-regime generalization. A recent study~\citep{Zhu_2021} showed that models trained on low-performance operational regimes may perform poorly in high-performance scenarios, underscoring the importance of regime adaptation. Their findings suggest that carefully aligning operational parameters across devices is necessary to achieve reliable generalization for next-generation burning-plasma tokamaks.

Together, these studies demonstrate the growing role of ML in disruption prediction and control. Early work established feasibility and cross-device generalization, subsequent efforts have moved toward real-time integration and interpretability. However, most of these work have been conducted as short term studies, as such, significant challenges remain, particularly in uncertainty quantification, adaptation to new operational regimes, continuous data drifts, and ensuring robustness for safety-critical deployment.

\subsection{Recent work on Uncertainty Quantification for ML}
\label{subsec:uq4ml}
Uncertainty quantification (UQ) has emerged as a crucial aspect of Machine Learning (ML) research, with applications in areas such as reliability engineering, risk assessment, and decision-making under uncertainty~\citep{Nemani2023uqml}. 
Previous studies have shown that several methods can be used to quantify uncertainty in ML models. These include Bayesian Neural Networks (BNNs), which incorporate prior distributions over model weights to provide probabilistic predictions~\citep{blundell2015weight}. A simpler approach is through Monte Carlo dropout (MCD), which uses dropout to approximate Bayesian neural networks \citep{gal2016dropout}. 
Another approach is through deep ensemble methods, which combine the predictions of multiple models to provide a more accurate prediction and uncertainty estimation~\citep{lakshminarayanan2017simple}.
Deep ensembles have been used for uncertainty quantification in ML-based energy confinement time extrapolation~\citep{nam2025machine}.
Although effective, Ensembles and BNNs require multiple models or repeated inference runs, which can be resource-intensive and impractical in environments with limited memory or requiring real-time processing. 
DQR~\citep{koenker_2005} based uncertainty quantification estimates prediction intervals by directly modeling conditional quantiles, capturing the inherent uncertainty in the data without requiring multiple inference calls or models. 
In summary, although these methods perform well for in-distribution uncertainty estimation (sometimes further enhanced through post-training calibration), they are not designed to provide reliable estimates under out-of-distribution conditions.

Recently, the DGPA technique has been introduced to approximation Gaussian Process (GP) kernel via random fourier features to a fixed sized matrix. The DGPA combines the distance awareness benefit of GP and highly expressive nature of DNNs. 
Being an approximation method with a fixed sized kernel matrix, it does not suffer with the scaling issues as seen with traditional GP.
The DGPA has been leveraged for reliable uncertainty quantification for both classification and regression tasks \citep{rajput2020uncertainty, PhysRevAccelBeams.26.044602}. These studies show that the DGPA is more reliable in uncertainty estimation on both IID and OOD data.
As such, we select the DGPA method for this study to produce reliable uncertainty estimation.



\subsection{Recent work related to online learning on non-stationary data streams}
Recent work on online learning for non-stationary data streams focuses on continuously updating models to quickly adapt to evolving data distributions caused by concept drift. Techniques such as partial or full model parameter updates based on the most recent data are explored to maintain prediction accuracy under abrupt, gradual, or incremental drift without emphasizing knowledge retention from the past \citep{IJCAI2022}. Proactive model adaptation frameworks estimate drift ahead of time and adjust model parameters accordingly to reduce the lag in adaptation caused by delayed ground-truth availability \citep{Zhao2024}. Lightweight ensemble methods and sliding window approaches offer efficient mechanisms to detect and respond to drift patterns in real-time data streams, allowing models to stay up to date with minimal computational overhead \citep{YangShami2021}. However, online learning is yet to be explored on Fusion applications.
\section{Methods}
\label{sec:methods}
We employ DNNs composed of convolutional blocks consisting of convolutional, maxpool, and activation layers. The output of the last convolutional block is flattened before feeding to dense blocks. Each dense block consist of dense, activation, and dropout layers. 
We leverage the DGPA layer as the final output layer.
The model is trained to produce expected coil deflection based on plasma input parameters in a supervised manner with Mean Absolute Error (MAE) loss and Adam optimizer with learning rate of $10^{-4}$. 
The model parameters are described in table~\ref{tab:model}.

\begin{table}[htbp]
\centering
\caption{Model Architecture and Hyperparameters, the entries in parenthesis represents the last block wherever it is different from the rest.}
\label{tab:model}
\begin{tabular}{ll}
\hline
\textbf{Component} & \textbf{Description} \\
\hline
Number of Conv1D blocks &  3 \\
Number of filters & 128 (64) \\
Kernel size & 3 \\
Conv stride & 1 \\
MaxPool1D Pool size & 2 \\
MaxPool1D Pool Strides & 2 \\
Number of Dense layers & 3 \\
Number of Dense Nodes & 128 (1) \\
Dropout rate & 0.05 \\
Activation & ReLU \\
Output Activation & linear \\
Loss function & MAE \\
Optimizer & Adam \\
Learning Rate & $10^{-4}$ \\

\hline
\end{tabular}
\end{table}

\subsection{Data Description and Pre-Processing}
Our study utilizes experimental data from the DIII-D fusion facility, which provides a unique opportunity to investigate the mechanical response of TF 
coils under electromagnetic load during plasma experiments.
A single experimental discharge is referred to as a \textit{shot}. Shots can vary in duration and are often used as independent instances for training and testing machine learning models.
In DIII-D, TF-coil deflection data consist of time-resolved measurements of coil motion during shots, recorded by displacement sensors (potentiometers) mounted on the 
outer coil bundles \citep{A24375_DIIID_TFDeflection}.
These diagnostics are synchronized with plasma and coil current signals, providing a direct representation of the mechanical response of the TF coils to 
Lorentz forces.

The data used in this study consist of nine plasma input variables, including B-coil current, E-coil current, input plasma current, and six magnetic 
field variables (pcf6a, pcf6b, pcf7a, pcf7b, pcf9a, and pcf9b).
These input variables are used to train the ML model, which is designed to produce one of 25 deflection values.
It is worth noting that all the deflection variables are highly correlated, and our model can be easily modified to predict all 25 deflection variables.

Prior to training the model, we perform pre-processing on the data. We remove any shots with missing values or not-a-number entries to ensure that the 
data is clean and consistent.
In some instances, the potentiometer can get stuck and produce noise measurements close to zero; these shots are removed to avoid contamination in model 
training.

To accommodate variable-sized time series data due to variations in shot lengths and to limit the model size, our model takes a small window (100 points) of input variables and 
produces deflection values for the last step in the window.
This approach allows the model to learn from the temporal trend in the recent window of the input data.
It is essential to note that the input to the model does not include the target variable to avoid initial seed bias.

\subsection{Online Learning}
\label{subsec:onlinelearning}
Online learning frameworks update ML models sequentially as new data become available, rather than retraining on the entire dataset, thereby enabling dynamic adaptation to evolving data distributions. This paradigm is particularly well-suited for our application, where the data stream exhibits non-stationary behavior. To accommodate the sequential nature of the data, we adopt a sliding-window batched training strategy in which the model is fine-tuned from its previously learned weights using the most recent sample together with a small buffer of recent historical data.
For abrupt drifts, shorter buffer size is favored, whereas for slow and gradual drifts, longer buffer size is more appropriate.
In fusion experiments, the drifts can attain different behaviors depending on maintenance schedules, experimental evolution and equipment wear-and-tear. As such, a single fixed buffer size for the online batched learning is not ideal. In this study we use an ensemble of models with different buffer sizes belonging to different time scales at DIII-D facility, starting with 1 shot (immediate scale), 5 shots (roughly an hour), 20 shots (about half a day), 40 shots (about a day worth of data), and 200 shots (a week full of data).

\subsection{Deep Gaussian Process Approximation}
Gaussian process (GP) \citep{NIPS1995_7cce53cf} models provide principled uncertainty quantification by leveraging kernel functions to assess similarity between samples, making them intrinsically distance-aware and effective at detecting OOD data. However, traditional GPs scale poorly with large datasets due to their high computational complexity, limiting their applicability to high-dimensional problems. 
To overcome these limitations, the DGPA model~\citep{PhysRevAccelBeams.26.044602} integrates the expressive capacity of DNNs with a fixed-size GP approximation. 
Specifically, it uses a Radial Basis Function (RBF) kernel approximation using Random Fourier Features (RFFs) at the output layer of the neural network as shown in equation~\ref{eq:rff}. 

\begin{equation}
    K(x, x') \approx \Phi(x)^\top \Phi(x'), \quad \text{with } \Phi \in \mathbb{R}^{D \times m}
    \label{eq:rff}
\end{equation}

where $m$ is the number of Fourier features (e.g., $m = 512$ in this study). This approach provides quality uncertainty estimation while mitigating scaling issues as with traditional GP.

To enforce distance-preservation between the input and hidden layers (ensuring that OOD samples can be reliably identified), a bi-Lipschitz constraint is applied to the transformation from input $x$ to the final hidden layer output $h(x)$:

\[
L_1 \cdot \|x_1 - x_2\| \leq \|h(x_1) - h(x_2)\| \leq L_2 \cdot \|x_1 - x_2\|
\]
with fixed constants $L_1 = 0.75$, $L_2 = 1.25$. The soft penalty on violations of this constraint is combined with the prediction loss (e.g., MAE) during training, promoting robust OOD detection. 
To ensure the uncertainty estimation is reliable on IID data points, we leverage the uncertainty toolbox \citep{chung2021uncertaintytoolboxopensourcelibrary} to calibrate the model with proper scaling of variance. The calibration routine is called at the end of each training session via tensorflow callbacks~\citep{tensorflow2015-whitepaper}.
This approach provides several benefits such as providing distance-aware uncertainty estimates, avoiding the computational cost of traditional GP by using RFF approximation, and providing uncertainty estimation without the need of multiple model inference calls. 

\begin{algorithm}[t]
\caption{Uncertainty-Aware Online Ensemble Learning for Non-Stationary Data Streams}
\KwIn{Ensemble size $n$; a base model ($\theta$) trained on data up to time $T$, initial buffers for each model $\mathcal{B}=\{\mathcal{B}_1, \mathcal{B}_2 \dots, \mathcal{B}_n\}$, learning rate $\eta$}
\KwOut{Ensemble predictions $\hat{y}^{(t)}$ over time}

\BlankLine
\textbf{Initialization:} Create an ensemble of models $\mathcal{E} = \{\theta_1, \theta_2 \dots, \theta_n\}$, initialized as $\theta_i = \theta$ 

\For{each time step $t = T+1, T+2, \dots$}{
    Obtain prediction and uncertainty estimation $\hat{y}_i^{{(t)}}$, $\sigma_i^{(t)} \gets f_{\theta_i}(x^{(t)})$ for $i=1, \dots, n$

    
    Compute ensemble prediction: $\hat{y}^{(t)} = \sum_{i=1}^n w_i^{(t)} \hat{y}_i^{(t)}$ with $w_i \propto \frac{1}{\sigma^2_i}, \text{and} \sum_{i=1}^n w_i^{(t)} = 1$
       
       Compute combined standard deviation $\sigma_{\bar{x}} = \left(\sum_{i=1}^n \sigma_i^{-2}\right)^{-\tfrac{1}{2}}$
    
    Receive new ground truth $\mathbf{y}^t$ from the experiment and update each rolling buffer $\mathcal{B}_i$ with $\mathbf{y}^t$; \\
    \For{each model $\theta_i$ and respective buffer $\mathcal{B}_i$}{
    Compute loss: $L = \frac{1}{|\mathcal{B}_i|} \sum_{(x^{(j)}, y^{(j)}) \in \mathcal{B}_i} \ell(f_{\theta_i}(x^{(j)}), y^{(j)})$ \\
         Backward pass: compute gradients $\nabla_{\theta_{i}} L$ \\
         Update parameters: $\theta_i \leftarrow \theta_i - \eta \nabla_{\theta_i} L$

    Perform post training uncertainty calibration as described in algorithm \ref{alg:uncertainty_calibration} in \ref{app:calibrationalgorithm}
    }
}
\label{alg:UAOL}
\end{algorithm}

\subsection{DGPA guided Online Learning}
To adapt to various drift types, we employ an ensemble of ML models trained on different buffer sizes of historical data in an incremental manner. However, relying solely on naive ensembles can lead to misleading results due to the lack of knowledge about expected error from individual models at the time of combining predictions.
To address this challenge, we propose leveraging uncertainty quantification from DGPA models as an estimate of the expected error per prediction. This enables guidance for meta-algorithms to generate more accurate ensemble predictions.
Calibrated uncertainties from the DGPA model can provide an estimate of error on predictions, where the predictions with lower uncertainties are expected to be more accurate than those associated with higher uncertainty values.
This is highlighted by figure \ref{fig:UQMAEComparison} that demonstrates a strong correlation between average model error and  uncertainty estimation from the DGPA model.
Our proposed uncertainty guided online ensemble method is outlined in Algorithm \ref{alg:UAOL}, which demonstrates how reliable uncertainty quantification can be leveraged to perform an informed weighted average of predictions from multiple models.
\section{Results}
\label{sec:results}

\begin{figure*}[t]
    \centering
    \begin{subfigure}[t]{0.4\textwidth}
        \centering
        \includegraphics[width=\textwidth]{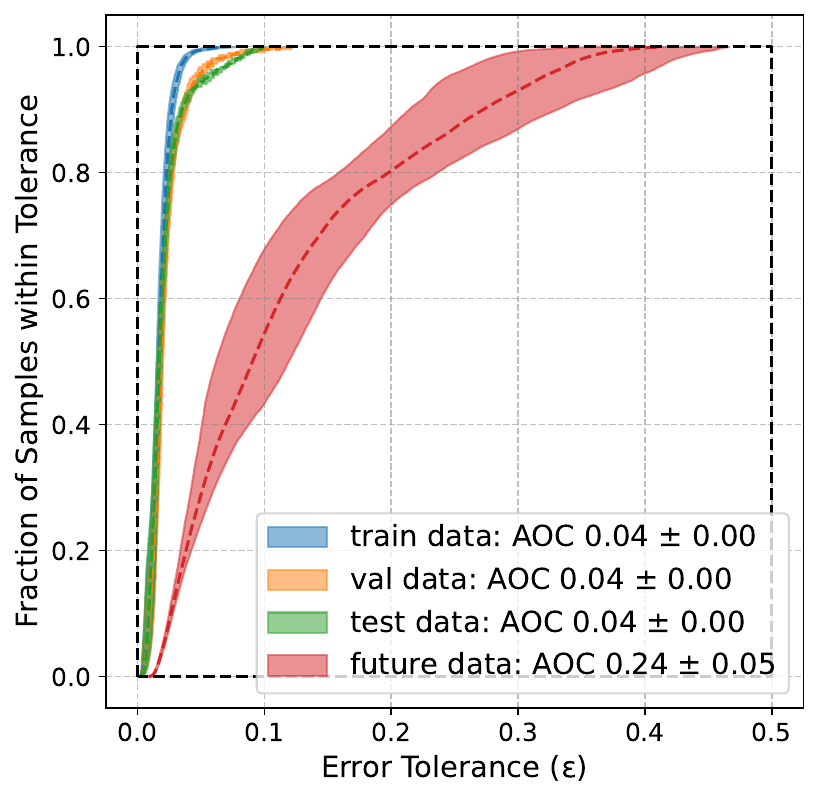}
        \caption{REC curves on train, validation and test data as well on on future data. The AOC is shown in the legends.}
        \label{subfig:rec_all}
    \end{subfigure}
    \begin{subfigure}[t]{0.4\textwidth}
        \centering
        \includegraphics[width=\textwidth]{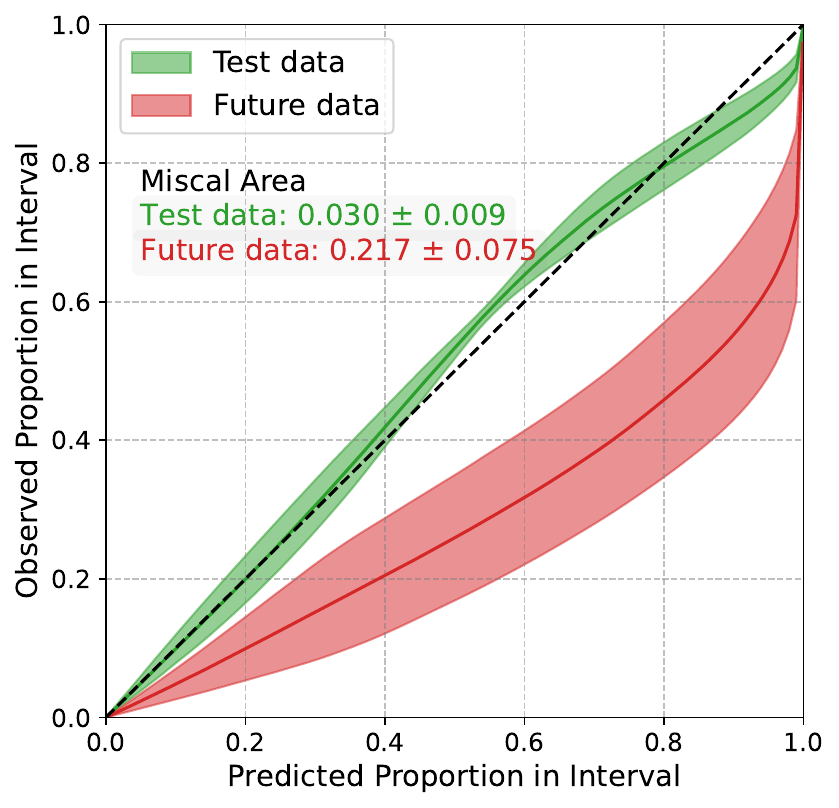}    
        \caption{Miscalibration area curves for test data from training distribution, and future data. The miscalibration area proportion is shown on the plot.}
        \label{subfig:Miscal}
    \end{subfigure}
    \caption{Performance of ML models on both OOD and future data. The bands represent 1-standard deviation spread in measurements over 10 trials.}
    \label{fig:threepanel}
\end{figure*}

\begin{figure*}[!t]
    \centering
    \begin{subfigure}[t]{0.45\textwidth}
        \centering
        \includegraphics[width=\textwidth]{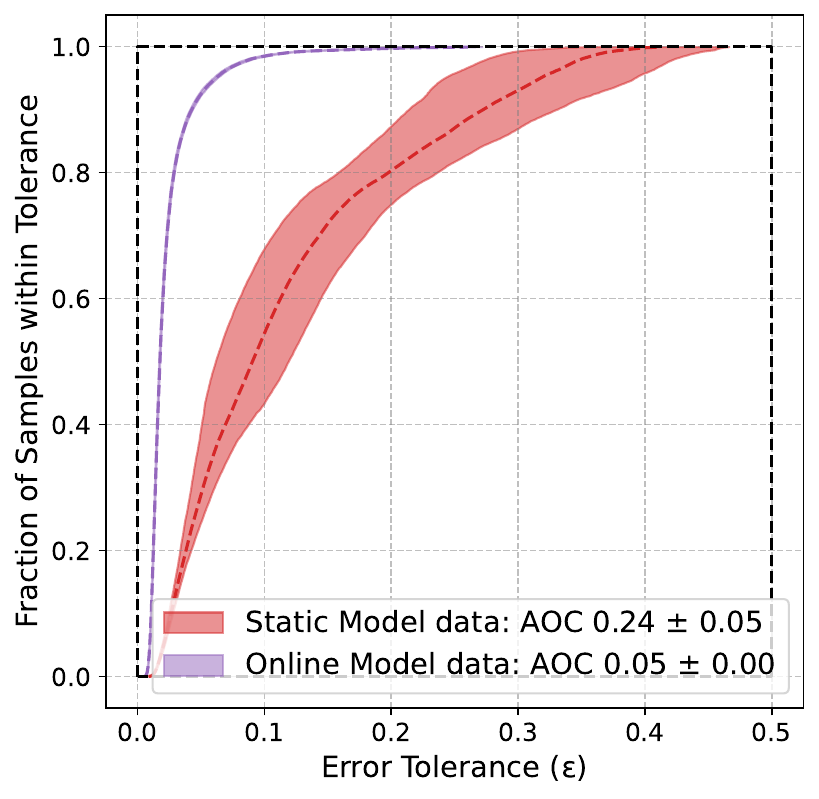}
        \caption{REC curves comparison between static models and online learning models on future data. The AOC is shown in the legends.}
        \label{subfig:rec_staticvsonline}
    \end{subfigure}
    \begin{subfigure}[t]{0.45\textwidth}
        \centering
        \includegraphics[width=\textwidth]{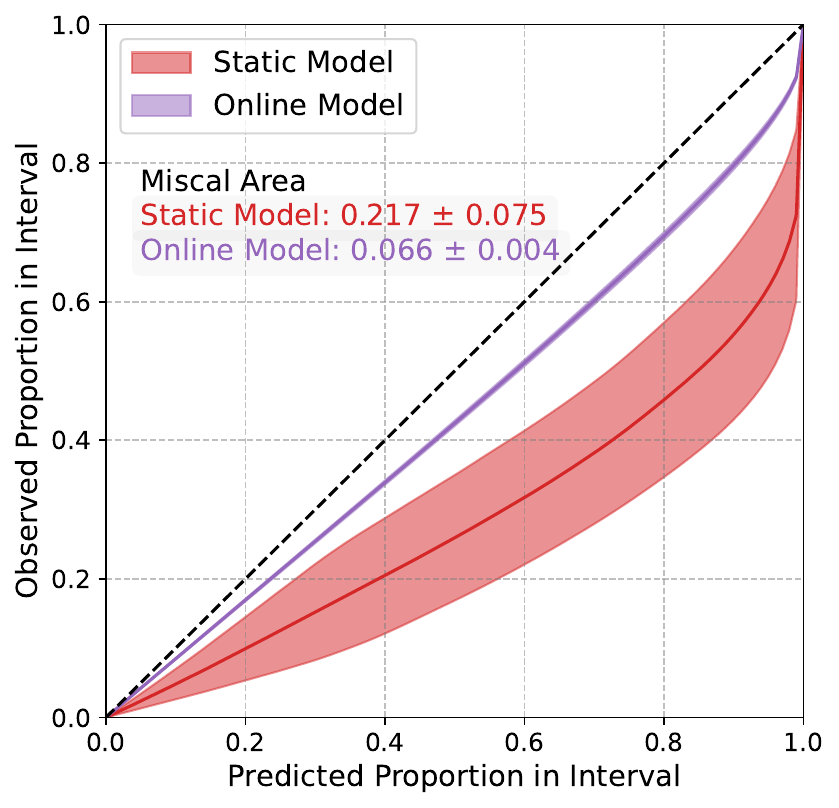}    
        \caption{Miscalibration area curves, the miscalibraton area proportion in shown on the plot.}
        \label{subfig:Miscal_staticvsonline}
    \end{subfigure}

    \vspace{0.5em}

    \begin{subfigure}[t]{0.95\textwidth}
        \centering
        \includegraphics[width=0.99\textwidth]{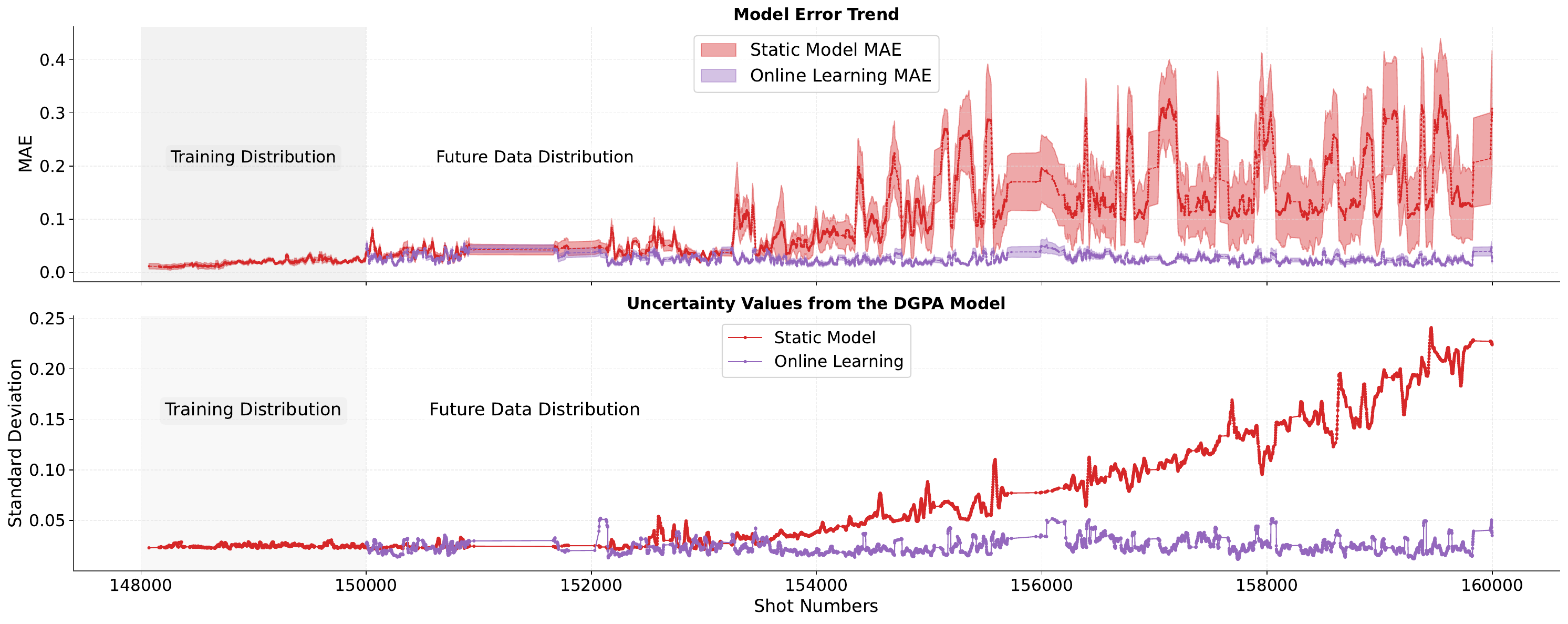}
        \caption{The top plot shows MAE per shot for both training data distribution (test dataset), and future data. The band shows 1-standard deviation spread over 10 trials. The model maintains performance close to the training data but starts to break down as the data drifts further. The bottom plot shows average uncertainty predicted by the DGPA model per shot. The uncertainty values correctly shows increasing trend as the data gets OOD.}
        \label{subfig:StaticModel}
    \end{subfigure}
    
    \caption{Comparison of Static and Online Learning model}
    \label{fig:staticvsonline}
\end{figure*}

\begin{figure*}
    \centering
    \begin{subfigure}[t]{0.95\textwidth}
        \centering
        \includegraphics[width=\textwidth]{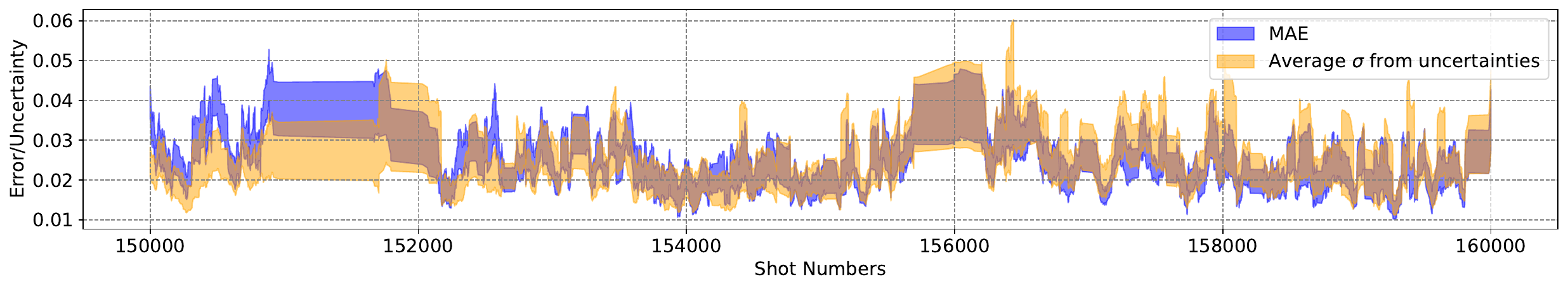}
        \caption{Overlaying model prediction error and 1-standard deviation based on uncertainty estimation from the DGPA model. The bands show spread from 10 independent model training with random initialization. The Pearson correlation coefficient between the mean of the two bands is 0.58.}
        \label{subfig:maevsuq}
    \end{subfigure}
    \begin{subfigure}[t]{0.95\textwidth}
        \centering
        \includegraphics[width=\textwidth]{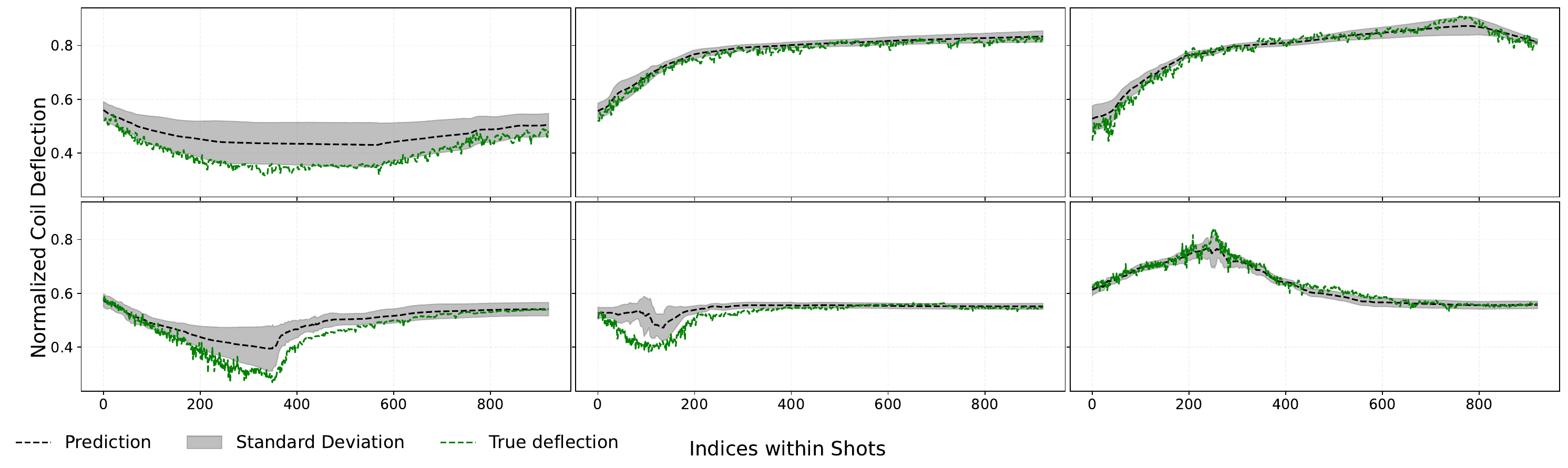}    
        \caption{Visualizing model predictions with DGPA uncertainty bands along with true target values. The DGPA uncertainty bands nicely correlated with the prediction error.}
        \label{subfig:overlay}
    \end{subfigure}
    \caption{Correlation between uncertainty estimation and prediction error from the DGPA model}
    \label{fig:UQMAEComparison}
\end{figure*}


\begin{figure*}
    \centering
    \includegraphics[width=0.99\textwidth]{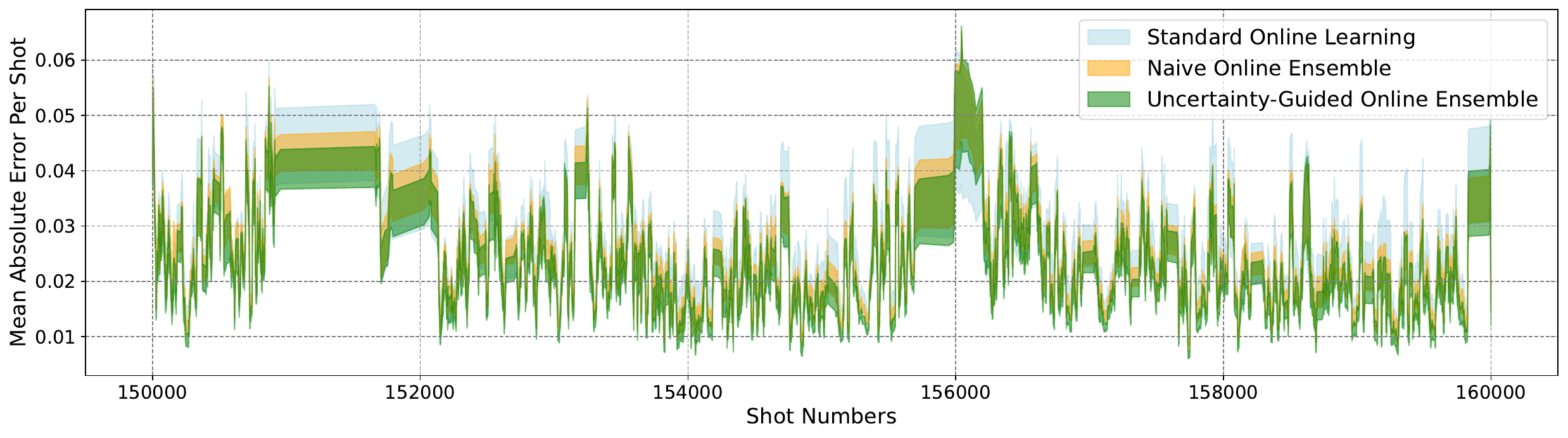}
    \caption{Comparison of Mean Absolute Error (MAE) per shot among different methods with 10 trials to produce statistically robust analysis. The figure shows moving average over 20 points to de-clutter the visualization and clearly show the trend. 
    }
    \label{fig:comparison}
\end{figure*}

\noindent
\textbf{Static Model:}
We conducted a thorough evaluation of our DNN model built with the architecture and hyper-parameters described in Section~\ref{sec:methods}.
The model is trained on the shot numbers 148,043 to 149,999. After pre-processing and removing noisy data samples, we obtained 1477 shots, which were divided 
into train, validation, and test datasets with a 70-15-15\% split ratio. Each shot consisted of a timeseries of length 1020, further segmented into 
windows of 100 points, resulting in 952,314 training samples, 203,541 validation samples, and 204,462 test samples.

To ensure statistical robustness, we performed 10 trials of model training with different random initialization and evaluated the models using a range of metrics, including Regression Error 
Characteristics (REC), Area over the REC Curve (AOC), Mean Absolute Error (MAE), Mean Squared-Error (MSE), and Mean Absolute Percentage Error (MAPE). We 
also assessed generalization to unseen data by evaluating the models on validation and test datasets, which were assumed to be independent and 
identically distributed (IID) with the training data.

The results of our evaluation are presented in Figure~\ref{subfig:rec_all}, which shows the REC curves for training, validation, test, and future data. 
Each band represents 1-standard deviation spread over 10 trials. The consistent behavior of the three curves for train, validation, and test data 
demonstrates strong model generalization to IID data, supported by the similar areas under the REC curve.

However, when evaluating the model on future data without any updates to its weights and biases (mimicking an online deployment scenario with no online updates), we observed a 
significant degradation in performance. As shown in Figure~\ref{subfig:rec_all}, the blue band representing future data exhibits poor performance, with a 
much larger REC area compared to the other datasets. This is further substantiated by the high AOC value of 0.24, compared to 0.04 for the other 
datasets.

To evaluate the quality of our uncertainty quantification (UQ) estimates, we employed the miscalibration area plots from the uncertainty toolbox~\citep{chung2021uncertaintytoolboxopensourcelibrary}. Figure~\ref{subfig:Miscal} presents the UQ miscalibration area plot, with bands showing 1-standard deviation spread over 10 
trials. The plot reveals that our model is well-calibrated on in-distribution data (test set), with a low calibration error of approximately 3\%. However, when 
evaluated on future data, the model exhibits significant overconfidence, as evident from the curves hanging below the ideal diagonal with a calibration 
error of about 22\%.

\noindent
\textbf{Online Learning:} 
To address the challenges posed by non-stationary data distributions, we investigated the efficacy of online learning and adaptation. Specifically, our 
model was incrementally trained on future data, incorporating one new shot at a time, beginning from shot number 150{,}000. Various buffer sizes for 
online batched training were evaluated as described in Section~\ref{subsec:onlinelearning}. Figure~\ref{fig:staticvsonline} presents a comparative 
analysis between the previously discussed static model and the incrementally updated online learning model.
For this comparison, we selected the best-performing model from the ensemble (batched training window size = 5). The
Figure~\ref{subfig:rec_staticvsonline} illustrate the REC curve comparison, which demonstrates that the online learning 
model substantially outperforms the static model. This improvement is expected, as the online model continuously adapts to the evolving data 
distribution. The AOC decreases from 0.24 to 0.05, approaching the model's performance on IID data.
Furthermore, Figure~\ref{subfig:Miscal_staticvsonline} presents a significant enhancement in uncertainty calibration, with miscalibration proportion 
reduced from approximately 22\% to 7\%, representing a \textbf{68\% improvement}. This improvement yields more reliable and well-calibrated uncertainty 
estimates.

Figure~\ref{subfig:StaticModel} presents a comparison of the mean absolute error (MAE; top) and average predictive uncertainty (bottom) between the 
static and online learning models. The static model exhibits an upward trend in prediction error as the data distribution drifts away from the training 
set. Although it becomes slightly overconfident on future data, its increased uncertainty estimates correctly indicate out-of-distribution (OOD) 
behavior.
In contrast, the online learning model maintains stable performance across future shots, achieving significantly lower MAE per shot and displaying 
well-calibrated confidence owing to its continual adaptation to new data. The overall MAE of the static model is about 0.119, whereas the online learning 
model achieves an MAE of 0.024 on future data. As such, the online learning model reduces the overall MAE by approximately \textbf{80\%}, underscoring 
the effectiveness of continual adaptation in addressing non-stationary data distributions.
The results suggest that online learning enables the model to adapt to changing data distributions, thereby improving both predictive performance, and uncertainty calibration.

\begin{table*}
\centering
\caption{Performance Comparison on the data from shot number 150,000 to 160,000. Static model with no online updates performs worse as expected due to data drifts. Among online learning methods, we select single model online learning as baseline, naive online ensemble and uncertainty guided online ensemble methods provide about 6\% and 10\% improvements respectively on MAE. The entries in table also shows associated 1-standard deviation error over 10 trials.
}
\label{tab:results}
\resizebox{\textwidth}{!}{
\begin{tabular}{lllll}
\hline
\textbf{Metric} & \textbf{Static Model} & \textbf{Online Learning}  & \textbf{Online Ensemble} & \textbf{UQ Guided OE}\\
\hline
MAE ($\times10^{-2}$) & 11.88$\pm$2.50 & 2.41$\pm$0.05 (baseline) & 2.26$\pm$0.02 (5.92\%)& \textbf{2.16$\pm$0.02 (10.24\%)}  \\
MSE ($\times10^{-2}$) & 2.57$\pm$1.11 & 0.14$\pm$0.00 (baseline) & 0.13$\pm$0.00 (8.66\%) & \textbf{0.12$\pm$0.00 (12.56\%)}  \\
MAPE & 22.44$\pm$4.61 & 4.18$\pm$0.10 (baseline) & 3.92$\pm$0.05 (6.30\%) & \textbf{3.80$\pm$0.04 (9.08\%)}  \\

\hline
\end{tabular}
}
\end{table*}

\noindent
\textbf{Online Ensembles:} 
To further improve performance, we implement two variations of online ensembles trained with batched incremental training using different buffer sizes as 
discussed in Section \ref{sec:methods}.
Firstly, we employ a naive ensemble that combines the predictions of several models by averaging them. Secondly, we leverage uncertainty estimation to 
perform an informed weighted average of the predictions as described in Section~\ref{sec:methods}.

Our proposed uncertainty guided online ensemble method is based on the assumption that the uncertainty estimation reliably indicates an approximation of the prediction errors, which is 
expected from a calibrated DGPA model.
Figure \ref{fig:UQMAEComparison} shows that this assumption holds for our application, where the uncertainty estimation and MAE from the DGPA models are 
strongly correlated (Pearson correlation coefficient of 0.58).
Figure \ref{subfig:maevsuq} overlays MAE and standard deviation 
estimated from the uncertainty predictions for an online learning model.
The bands in this figure show the spread across 10 trials to provide a statistically robust comparison.
Figure \ref{subfig:overlay} shows model predictions with uncertainty bands (here the bands are based on DGPA uncertainty predictions), along with the 
true target values. Together, these two figures clearly demonstrate that the model uncertainties are strongly correlated with the prediction error.

Next, we show a comparison among standard online learning method (we pick the best performing single model among the ensemble of models), naive online 
ensemble, and our proposed uncertainty guided online ensemble methods in Figure \ref{fig:comparison}.
For both ensembles, the different models are trained with different buffer sizes for batched incremental training but kept consistent between the two 
methods.
To provide a statistically robust comparison, we perform 10 trials of each method. The bands in the figure represent statistical variations over multiple 
trials.

It is evident that the uncertainty guided online ensemble method outperforms naive ensemble and a single model approach on almost all the shots.
The figure indicates that the ensemble method improves the performance over standard single model approach, while uncertainty guidance further improves 
the performance.
This result is further supported by different error metrics as shown in Table \ref{tab:results}.
Considering standard single model online learning as baseline, naive ensemble achieves 5.92\% better MAE, whereas uncertainty guided online ensemble 
achieves about 10.24\% better performance in terms of MAE.

The REC curvers among the online learning methods including the two ensemble variations are similar. The uncertainty calibration quality also holds when the predictions and uncertainty estimation are combined for the ensemble methods.
The comparison of REC and Calibration quality among the online learning methods are provided in Appendix \ref{appendix:a}.

\textbf{Discussion:} 
Although the uncertainty-guided online ensemble outperforms the standard single-model online learning approach, it requires substantially greater computational and memory resources to concurrently train and perform inference with multiple models. As such, there is a trade-off between predictive performance and computational efficiency. Online learning remains a critical capability for adapting to evolving data distributions; however, ensemble-based extensions may be more appropriate in scenarios where additional performance improvements are strongly desired and computational resources permit.

In this work, the ensemble models were trained using different buffer sizes of historical data in a batched incremental fashion, primarily to address various types of distributional drifts (e.g., abrupt and gradual). Nonetheless, the proposed framework is broadly applicable to other domains where ensemble members may comprise heterogeneous model architectures designed to capture different data sensitivities.

For uncertainty estimation, this study employed the Deep Gaussian Process Approximation (DGPA) method, owing to its ability to produce reliable out-of-distribution (OOD) estimates and well-calibrated in-distribution uncertainties (with post-training calibration) for error estimation. However, the ensemble framework is not limited to DGPA; any uncertainty estimation technique demonstrating a strong correlation between predicted uncertainty and actual error can be integrated to guide the ensemble prediction mechanism.
\section{Conclusion}

To bridge the gap between machine learning (ML) developments and long-term deployment in fusion applications due to non-stationary data streams, we have presented an application of online learning to predict TF-coil deflection at the DIII-D Tokamak.
Our approach leverages the DGPA model to provide reliable uncertainty estimation alongside predictions, which is critical for ensuring whether the model's predictions are trustworthy.
Our study demonstrates that online learning reduces the prediction error by \textbf{80\%} and improve the uncertainty calibration of the DGPA model by about \textbf{68\%}.
We have explored naive online ensembles and proposed a novel uncertainty-guided online ensemble method that leverages uncertainty estimation to intelligently combine predictions from multiple models.
The results show that the proposed uncertainty-guided online ensemble method outperforms standard online learning, achieving about \textbf{10\%} reduction in mean absolute error (MAE).
Our method with different buffer sizes for ensembles is particularly suitable for non-stationary online learning scenarios where drift is expected at various time scales. 
However, it is essential to note that the algorithm can be generalized to other applications that may benefit from different types of learners.

In future, we aim to further improve uncertainty quantification (UQ) based error estimation and extend our approach to include other types of models in the ensemble, including different architectures. 
We also plan to investigate the use of attention mechanisms with various scales to address drifts.
Moreover, we intend to deploy this approach at the DIII-D facility to evaluate its performance in real-time.
By addressing these challenges and expanding on our approach, we hope to contribute significantly to the development of reliable and robust ML models for 
fusion applications.

\section{Acknowledgment}
Work supported by Jefferson Lab and General Atomics internal research and development. 
This manuscript has been authored by Jefferson Science Associates (JSA) operating the Thomas Jefferson National Accelerator Facility for the U.S.Department of Energy under Contract No. DE-AC05-06OR23177. 
Additionally, this material is based upon work supported by the U.S. Department of Energy, Office of Science, Office of Fusion Energy Sciences, using the DIII-D National Fusion Facility, a DOE Office of Science user facility, under Award(s) DE-FC02-04ER54698, along with Office of Fusion Energy Sciences Award No. DE-SC0024426 (Fusion Data Platform).
The US government retains, and the publisher, by accepting the article for publication, acknowledges that the US government retains a nonexclusive, paid-up, irrevocable, worldwide license to publish or reproduce the published form of this manuscript, or allow others to do so, for US government purposes. 

\textit{
Disclaimer: This report was prepared as an account of work sponsored by an agency of the United States Government. Neither the United States Government nor any agency thereof, nor any of their employees, makes any warranty, express or implied, or assumes any legal liability or responsibility for the accuracy, completeness, or usefulness of any information, apparatus, product, or process disclosed, or represents that its use would not infringe privately owned rights. Reference herein to any specific commercial product, process, or service by trade name, trademark, manufacturer, or otherwise does not necessarily constitute or imply its endorsement, recommendation, or favoring by the United States Government or any agency thereof. The views and opinions of authors expressed herein do not necessarily state or reflect those of the United States Government or any agency thereof.
}

\appendix
\newpage
\section{REC and Calibration quality of online ensembles}
\label{appendix:a}

\begin{figure}[H]
    \centering
    \includegraphics[width=0.5\linewidth]{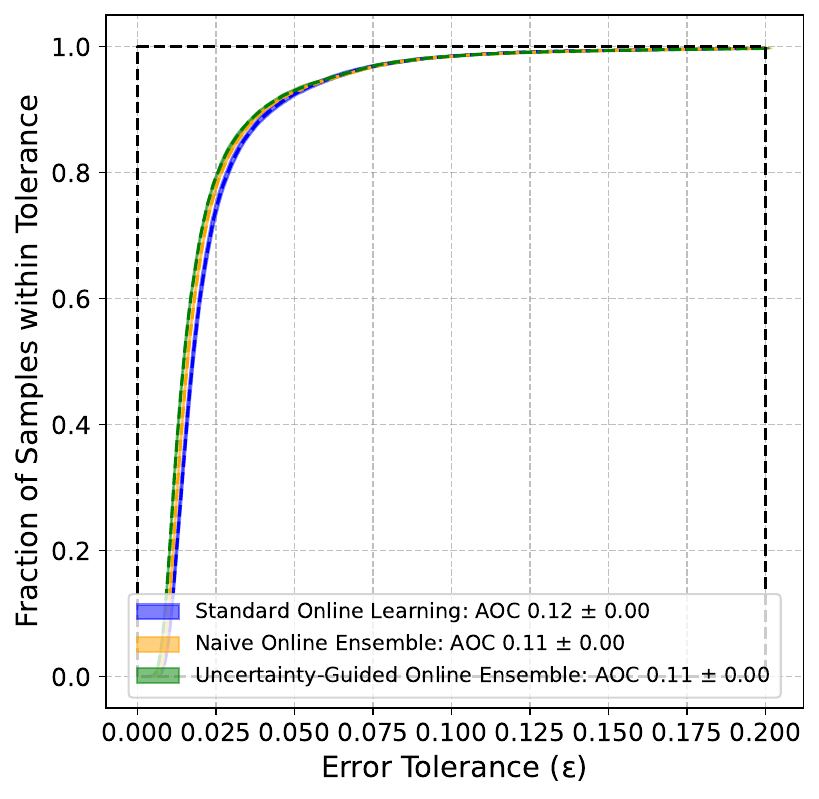}
    \caption{Comparison of REC among online methods. The REC for naive online ensemble and uncertainty guided online ensemble are slightly better than the standard single model online learning model. This is also reflected in AOC shown in the legends.}
    \label{fig:Ensemble_REC_Comparison}
\end{figure}

\begin{figure}[H]
    \centering
    \includegraphics[width=0.5\linewidth]{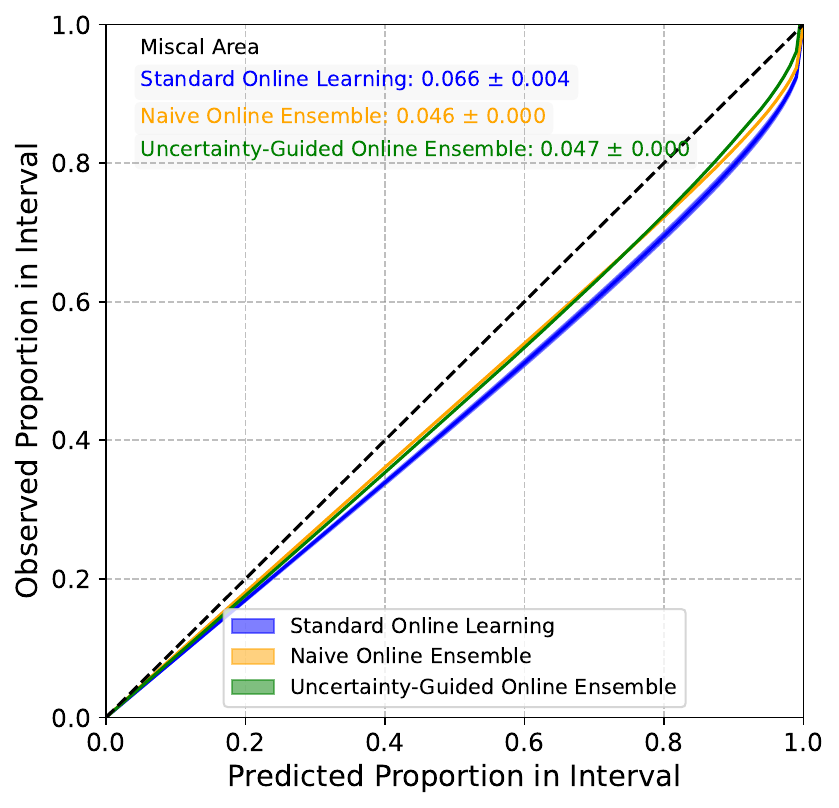}
    \caption{Miscalibration plot comparison among online methods. The online ensemble and uncertainty guided online ensemble both provide slight improvement in calibration compared to standard single model online learning. The bands represent statistical spread from 10 trials.}
    \label{fig:placeholder}
\end{figure}

\section{Post Training Uncertainty Calibration Algorithm}
\label{app:calibrationalgorithm}

\begin{algorithm}[H]
\caption{Uncertainty Calibration via Miscalibration Minimization}
\label{alg:uncertainty_calibration}

\KwIn{Predicted means $\{\hat{y}_i\}_{i=1}^N$, predicted stds $\{\sigma_i\}_{i=1}^N$, true targets $\{y_i\}_{i=1}^N$, initial uncertainty scaling $\alpha = 1.0$}
\KwOut{Scaling factor for calibrated uncertainty values}

\BlankLine
Define nominal coverage levels $\mathcal{P} = \{p_1, p_2, \dots, p_K\}$, $p_k \in [0,1]$ \\
\For{each $p_k \in \mathcal{P}$}
{
    Compute $z_{p_k} = \Phi^{-1}\!\left( \frac{1 + p_k}{2} \right)$
    Compute empirical coverage:
    \[
    \widehat{C}(p_k; \alpha)
    = \frac{1}{N} \sum_{i=1}^N 
      \mathbf{1}\!\left(
        |y_i - \hat{y}_i| \le 
        z_{p_k} \, (\alpha \sigma_i)
      \right)
    \]
}
Compute miscalibration loss:
\[
\mathcal{L}_{\text{miscal}}(\alpha)
= \frac{1}{M} \sum_{k=1}^{M}
  \big| \widehat{C}(p_k; \alpha) - p_k \big|
\]
Optimize calibration parameter(s):
\[
\alpha^{*} = \arg\min_{\alpha > 0} 
\mathcal{L}_{\text{miscal}}(\alpha)
\]
\textbf{Return:} $\alpha^{*}$

\end{algorithm}

\newpage
\bibliography{references}
\end{document}